\documentclass[runningheads]{llncs}
\usepackage[T1]{fontenc}
\usepackage{graphicx,verbatim}
\usepackage{amsmath,amssymb,hyperref,multirow}
\usepackage[table]{xcolor}
\begin{document}
\title{Hybrid Transformer-Mamba for Weakly Supervised Volumetric Medical Segmentation}
%

\author{Yiheng~Lyu\inst{1,2}\thanks{Corresponding author.} \and
Lian~Xu\inst{1} \and
Coen~Arrow\inst{1,2,3} \and
Mohammed~Bennamoun\inst{1} \and
Farid~Boussaid\inst{1} \and
Girish~Dwivedi\inst{1,2,4,5}}
\authorrunning{Y. Lyu et al.}
\institute{University of Western Australia, Crawley WA 6009, Australia
\email{yiheng.lyu@uwa.edu.au} \and
Harry Perkins Institute of Medical Research, Murdoch WA 6150, Australia \and
National Imaging Facility, Nedlands WA 6009, Australia \and
Fiona Stanley Hospital, Murdoch WA 6150, Australia \and
Victor Chang Cardiac Research Institute, Darlinghurst NSW 2010, Australia}
  
\maketitle              
\begin{abstract}
Weakly supervised segmentation enables model training from plane-level labels. Existing methods often rely on 2D encoders, neglecting the volumetric nature of medical data. We propose TranSamba, a hybrid Transformer-Mamba architecture designed to capture 3D context via cross-plane modeling. TranSamba augments a Vision Transformer backbone with Cross-Plane Mamba blocks, leveraging linear-time modeling for efficient information exchange across neighboring planes. This exchange improves in-plane self-attention and subsequent attention maps for object localization. TranSamba maintains linear time complexity and constant space complexity with respect to the input volume depth. Extensive experiments on three datasets covering diverse modalities and pathologies show that TranSamba achieves state-of-the-art performance, demonstrating the generalizable efficacy of cross-plane modeling. Code is available at: \url{https://github.com/YihengLyu/TranSamba}.


\end{abstract}
\section{Introduction}
Segmentation models for volumetric medical images such as computed tomography (CT) and magnetic resonance imaging (MRI) are typically trained from voxel-level labels, which are laborious to obtain. Weakly supervised segmentation trains models from labor-efficient slice-level labels \cite{chen2023ame,fu2023cam,lyu2024importance,schmidt2024tonno}, which contain no localization information. Class activation mapping (CAM) \cite{zhou2016learning} extracted from trained classification models can be used for coarse object localization. However, domain-specific characteristics introduce challenges for CAM-based methods. Existing strategies to tackle these challenges can be broadly categorized as: \textbf{1.} multi-scale feature aggregation \cite{chen2023ame,fu2023cam,ma2020ms}; \textbf{2.} addressing boundary and size characteristics \cite{chang2021weakly,chen2022c,lyu2024importance,yang2024anomaly}; \textbf{3.} self-supervised regularization \cite{kuang2023cluster,lyu2024importance,patel2022weakly,tang2021m}; \textbf{4.} incorporating target-specific constraints \cite{chen2023ws,chen2022c,dhamale2025inter,li2022deep}.

\begin{figure}[t]
\centering
\includegraphics[width=0.9\textwidth]{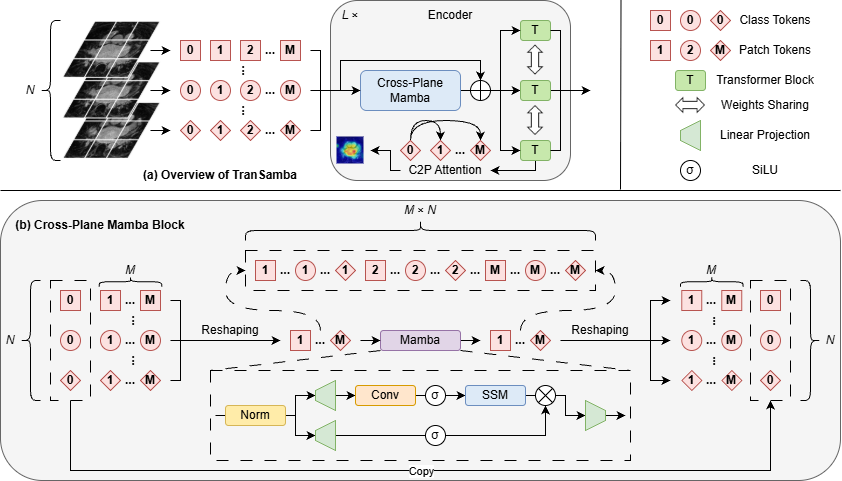}
\caption{\textbf{(a)} The encoder comprises $L$ hybrid layers of CPM and Transformer blocks. An input volume contains $N$ planes of $1 + M$ tokens. The $N$ sequences are reshaped into a single sequence for cross-plane modeling with Mamba and reshaped to $N$ sequences for parallelized in-plane SA. For inference, attention maps are generated from the C2P attention of the $L$ Transformer blocks. \textbf{(b)} The $N \times M$ patch tokens are separated from class tokens and reshaped into a single sequence of length $M \times N$, where neighboring tokens are from different planes. Cross-plane modeling is achieved by learning interactions between the $M \times N$ tokens via Mamba. The Mamba output is reshaped into $N$ sequences and concatenated with class tokens.} \label{fig1}
\end{figure}

The volumetric nature remains underexplored when training segmentation models from slice-level labels. 2D encoders are typically used to make predictions from individual slices that align directly with their labels. The Vision Transformer (ViT) \cite{dosovitskiy2020image} is a popular 2D encoder; its class-to-patch (C2P) attention, the pairwise self-attention (SA) \cite{vaswani2017attention} between class and patch tokens, is effective for weakly supervised segmentation \cite{xu2022multi,xu2024mctformer+}. We term C2P attention within a single slice as \textbf{in-plane} SA, which is complemented by \textbf{cross-plane} interactions across neighboring slices to capture the volumetric nature. However, cross-plane SA is computationally expensive, scaling quadratically with $N$ planes.

Built on structured state space models (SSMs), the Mamba architecture \cite{gu2024mamba} achieves linear-time scaling with sequence length. Mamba-based vision backbones \cite{liu2024vmamba,zhu2024vision} have emerged for computational efficiency; in medical segmentation, Mamba applications have succeeded under fully \cite{ma2024u,xing2024segmamba} and weakly supervised \cite{fan2024pathmamba,pan2025mamba} settings. Furthermore, Mamba-Transformer hybrids \cite{hatamizadeh2025mambavision,lieber2024jamba} are proposed alongside a novel pre-training strategy \cite{liu2025map} for hybrid architectures.

To leverage linear-time cross-plane modeling via Mamba and the efficacy of C2P attention, we propose TranSamba, a hybrid \textbf{TranS}former-M\textbf{amba} architecture for weakly supervised volumetric medical segmentation. We place a novel Cross-Plane Mamba (CPM) block before the Transformer block in every layer of a ViT encoder. The CPM blocks allow efficient information exchange between patch tokens across neighboring planes, enriching information encoding for each token; the Transformer blocks produce C2P attention maps via in-plane SA, which benefits from this enrichment. Without any of the aforementioned strategies, TranSamba achieves improved object localization via straightforward model training. Our \textbf{contributions} are: \textbf{1.} We propose TranSamba, a hybrid architecture combining Transformer for effective in-plane SA with Mamba for efficient cross-plane modeling. \textbf{2.} The time complexity of TranSamba scales linearly with the number of planes; for batch processing, its space complexity remains constant with the number of planes. \textbf{3.} TranSamba achieves state-of-the-art (SOTA) performance on three distinct datasets, underscoring the generalizable effectiveness of cross-plane modeling for weakly supervised volumetric medical segmentation.

\section{TranSamba}
Fig.~\ref{fig1}(a) shows the TranSamba overview. The encoder comprises $L$ hybrid layers, each consisting of a CPM block in series with a Transformer block. An input volume comprises $N$ planes, each transformed into a sequence of tokens $\mathbf{T}_{0}^{'} \in \mathbb{R}^{1 \times (1 + M) \times D}$ using the ViT image-to-sequence transformation; $M$ and $D$ denote the patch token number and the embedding dimension, respectively. In the $l$-th layer, the CPM input is obtained by stacking the $(l - 1)$-th layer output sequences, and cross-plane modeling is achieved with the CPM block:
\begin{equation}
\mathbf{V}_{l} = [[\mathbf{T}_{l-1, 1}^{'}], \dots, [\mathbf{T}_{l-1, N}^{'}]]
\end{equation}
\begin{equation}
\mathbf{V}_{l}^{'} = \text{CPM} (\mathbf{V}_{l}) + \mathbf{V}_{l}
\end{equation}
The CPM output is reshaped back to $N$ sequences for parallelized in-plane SA:
\begin{equation}
\mathbf{T}_{l, n} = \mathbf{V}_{l}^{'} [n, :, :], \quad n=1, \dots, N
\end{equation}
\begin{equation}
\mathbf{T}_{l, n}^{'}, \mathbf{A}_{l, n} = \text{Transformer} (\mathbf{T}_{l, n}), \quad n=1, \dots, N
\end{equation}
where $\mathbf{A}_{l, n} \in \mathbb{R}^{(1 + M) \times (1 + M)}$ denotes the pairwise SA.

During training, an output sequence $\mathbf{T}_{L}^{'}$ comprises a class token $\mathbf{T}_{L}^{class} \in \mathbb{R}^{1 \times D}$ and $M$ patch tokens $\mathbf{T}_{L}^{patch} \in \mathbb{R}^{M \times D}$. A global average pooling (GAP) layer processes $\mathbf{T}_{L}^{class}$: $\hat{y}^{class} = \text{GAP} (\mathbf{T}_{L}^{class})$; a convolutional layer followed by a global weighted ranking pooling (GWRP) \cite{kolesnikov2016seed} layer processes $\mathbf{T}_{L}^{patch}$: $\hat{y}^{patch} = \text{GWRP} (\text{Conv} (\mathbf{T}_{L}^{patch}))$. The training loss sums two cross-entropy (CE) terms: $\mathcal{L} = \text{CE} (\hat{y}^{class}, y) + \text{CE} (\hat{y}^{patch}, y)$. During inference, attention maps are generated by aggregating the C2P attention from all $L$ Transformer blocks:
\begin{equation}
\mathbf{A}^{\text{C2P}} = \sum_{l=1}^{L} \mathbf{A}_{l} [1, 2:]
\end{equation}

\begin{table}[t]
\centering
\caption{Computational complexity comparison. Shaded cells represent TranSamba.}\label{tab1}
\fontsize{8}{9.5}\selectfont
\begin{tabular}{ccc}
\hline
 & In-Plane & Cross-Plane\\
\hline
Sequence Length & $M$ & $M N$\\
\hline
\multicolumn{3}{c}{\textbf{Time Complexity}}\\
\hline
SA & \cellcolor{gray!50} $4 M D^2 + 2 M^2 D$ & $4 M N D^2 + 2 M^2 N^2 D$\\
SSM & $128 M D$ & \cellcolor{gray!50} $128 M N D$\\
\hline
\multicolumn{3}{c}{\textbf{Space Complexity}}\\
\hline
Batch Size & $B$ & $\frac{B}{N}$\\
SA & \cellcolor{gray!50} $B (4 M D^2 + 2 M^2 D)$ & $\frac{B}{N} (4 M N D^2 + 2 M^2 N^2 D)$\\
SSM & $B (128 M D)$ & \cellcolor{gray!50} $\frac{B}{N} (128 M N D)$\\
\hline
\end{tabular}
\end{table}

\subsection{Cross-Plane Mamba} \label{CPM}
Fig.~\ref{fig1}(b) shows a CPM block comprising a Mamba \cite{gu2024mamba} block and two reshaping operations. The CPM input $\mathbf{V} \in \mathbb{R}^{N \times (1 + M) \times D}$ comprises $N$ sequences, each containing one class token for global plane representation and $M$ patch tokens representing local patches. Cross-plane modeling with Mamba is achieved by efficiently learning interactions between the $M \times N$ patch tokens in linear time; cross-plane SA significantly increases computational complexity (Sec.~\ref{CCA}).

The $N$ class tokens are separated and excluded from the Mamba block. The $N \times M$ patch tokens are reshaped into a single sequence $\mathbf{x} \in \mathbb{R}^{(M \times N) \times D}$; neighboring patch tokens are from different planes, thereby prioritizing their interactions in cross-plane modeling with Mamba: $\mathbf{x}^{'} = \text{Mamba} (\mathbf{x})$. The Mamba output sequence $\mathbf{x}^{'}$ is reshaped into $N$ parallel sequences of $M$ patch tokens, each concatenated with its class token.

\subsection{Computational Complexity Analysis} \label{CCA}
Given a sequence $\mathbf{T} \in \mathbb{R}^{1 \times L \times D}$, the time complexity of global SA is $4 L D^2 + 2 L^2 D$; with default expanded state dimension ($2D$) and fixed parameter (16), the time complexity of global SSM is $128 L D$ \cite{zhu2024vision}. For a volume with $N$ planes, Tab.~\ref{tab1} summarizes the effective sequence length, batch size, and computational complexity of in-plane and cross-plane modeling. With the series arrangement of CPM and Transformer blocks, the time complexity of TranSamba is $128 M N D + 4 M D^2 + 2 M^2 D$ and scales linearly with $N$. Replacing SSM with SA for cross-plane modeling increases the time complexity to $4 M N D^2 + 2 M^2 N^2 D + 4 M D^2 + 2 M^2 D$, scaling quadratically with $N$. For a batch of $\frac{B}{N}$ volumes each with $N$ planes, the effective batch sizes for in-plane and cross-plane modeling are $B$ and $\frac{B}{N}$. The space complexity of TranSamba is $B (4 M D^2 + 2 M^2 D) + \frac{B}{N} (128 M N D)$, independent of $N$. Replacing SSM with SA for cross-plane modeling increases the space complexity to $B (4 M D^2 + 2 M^2 D) + \frac{B}{N}(4 M N D^2 + 2 M^2 N^2 D)$, which grows with $N$.

\begin{table}[t]
\centering
\caption{Ablation studies. Shaded rows indicate the default configuration of TranSamba (V3, 16 planes, CrossIn). CPM: cross-plane Mamba; IPM: in-plane Mamba; IPT: in-plane Transformer; Uni-: unidirectional; Bi-: bidirectional.}\label{tab2}
\fontsize{8}{9.5}\selectfont
\begin{tabular}{c|ccc|ccc}
\multicolumn{7}{c}{\textbf{(a)} Complexity comparison.}\\
\hline
Variant & CPM & IPM & IPT & Speed (FPS) & Memory (GB) & \#params. (M)\\
\hline
V1 & - & - & \checkmark & 3401.4 & 13.0 & 21.7\\
V2 & - & Uni- & \checkmark & 1764.4 & 17.6 & 33.3\\
\rowcolor{gray!50} V3 & \checkmark & - & \checkmark & 1777.7 & 17.3 & 33.3\\
V4 & - & Bi- & - & 1613.1 & 9.2 & 23.6\\
V5 & \checkmark & Bi- & - & 1126.6 & 13.5 & 35.2\\
\hline
\multicolumn{7}{c}{}\\
\end{tabular}
\begin{tabular}{c|cc|cc|cc}
\multicolumn{7}{c}{\textbf{(b)} Effectiveness of cross-plane modeling.}\\
\hline
\multirow{2}{*}{Variant} & \multicolumn{2}{c|}{BraTS} & \multicolumn{2}{c|}{KiTS} & \multicolumn{2}{c}{LASC}\\
 & DSC (\%) & IoU (\%) & DSC (\%) & IoU (\%) & DSC (\%) & IoU (\%)\\
\hline
V1 & 40.7 & 28.8 & 15.8 & 20.1 & 33.5 & 20.4\\
V2 & 41.9 & 28.2 & 19.5 & 24.7 & 38.1 & 24.3\\
\rowcolor{gray!50} V3 & \textbf{60.7} & \textbf{44.2} & \textbf{25.1} & \textbf{24.9} & \textbf{45.9} & \textbf{29.3}\\
V4 & 32.6 & 18.3 & 5.8 & 5.3 & 0.3 & 0.1\\
V5 & 36.7 & 19.5 & 6.0 & 5.5 & 1.4 & 0.6\\
\hline
\multicolumn{7}{c}{}\\
\multicolumn{7}{c}{\textbf{(c)} Number of planes.}\\
\hline
\multirow{2}{*}{\#planes} & \multicolumn{2}{c|}{BraTS} & \multicolumn{2}{c|}{KiTS} & \multicolumn{2}{c}{LASC}\\
 & DSC (\%) & IoU (\%) & DSC (\%) & IoU (\%) & DSC (\%) & IoU (\%)\\
\hline
128 & 57.6 & 43.6 & N/A & N/A & N/A & N/A\\
64 & 55.1 & 41.8 & 20.0 & 17.4 & N/A & N/A\\
32 & 53.7 & 40.8 & 22.3 & \textbf{27.7} & 40.8 & 24.9\\
\rowcolor{gray!50} 16 & \textbf{60.7} & \textbf{44.2} & \textbf{25.1} & 24.9 & \textbf{45.9} & \textbf{29.3}\\
\hline
\multicolumn{7}{c}{}\\
\multicolumn{7}{c}{\textbf{(d)} Hybrid layer design.}\\
\hline
\multirow{2}{*}{Design} & \multicolumn{2}{c|}{BraTS} & \multicolumn{2}{c|}{KiTS} & \multicolumn{2}{c}{LASC}\\
 & DSC (\%) & IoU (\%) & DSC (\%) & IoU (\%) & DSC (\%) & IoU (\%)\\
\hline
Parallel & 60.8 & 42.0 & 23.9 & 24.4 & \textbf{47.6} & \textbf{30.9}\\
InCross & \textbf{64.4} & \textbf{44.2} & 24.6 & \textbf{25.2} & 42.0 & 26.2\\
\rowcolor{gray!50} CrossIn & 60.7 & \textbf{44.2} & \textbf{25.1} & 24.9 & 45.9 & 29.3\\
\hline
\end{tabular}
\end{table}

\section{Experiments}
We formulate the task as binary segmentation; while important, multi-class segmentation requires techniques orthogonal to architectural novelty, such as modeling co-occurrence \cite{chen2022c} or inter-class \cite{chen2023ws,dhamale2025inter,xu2022multi,xu2024mctformer+} or inter-channel \cite{kuang2023cluster} correlations.

\textbf{Datasets} TranSamba is evaluated on three datasets. \textbf{1. Brain Tumor Segmentation (BraTS)} \cite{bakas2017advancing,bakas2018identifying,menze2014multimodal} in the Medical Segmentation Decathlon \cite{antonelli2022medical,simpson2019large}: targeting the whole tumor using the T2 fluid-attenuated inversion recovery sequence; \textbf{2. Kidney Tumor Segmentation (KiTS)} 2023 \cite{heller2021state}: targeting the kidney tumor; \textbf{3. Left Atrium Segmentation Challenge (LASC)} \cite{xiong2021global}: targeting the LA cavity. BraTS, KiTS, and LASC comprise 484, 489, and 154 volumetric studies, respectively; 100 studies per dataset are hidden for testing; the remaining studies use an 80/20 training/validation split.

\textbf{Evaluation} The ablation studies and SOTA comparisons are conducted on our validation and testing sets, respectively. The 3D mean Dice similarity coefficient (DSC) evaluates overall study-level segmentation performance; the 2D mean intersection-over-union (IoU) evaluates plane-level localization map quality. The 95\% Hausdorff distance (HD95) is sensitive to outliers under weakly supervised settings and is only used for variants with close DSC/IoU.

\textbf{Implementation} Our ViT backbone is built on DeiT-S \cite{touvron2021training}; CPM blocks employ vanilla Mamba \cite{gu2024mamba} blocks. For training, a batch comprises 16 randomly sampled volumes, each with 16 contiguous planes; all variants are trained for 100 epochs using only plane-level labels. We use the AdamW optimizer (initial learning rate $5 \times 10^{-4}$, weight decay 0.05) and a cosine decay schedule with a 5-epoch warmup. For inference, attention maps are generated in a single stage. Training and inference use a single 40GB NVIDIA A100 GPU.

\begin{figure}[t]
\centering
\includegraphics[width=0.9\textwidth]{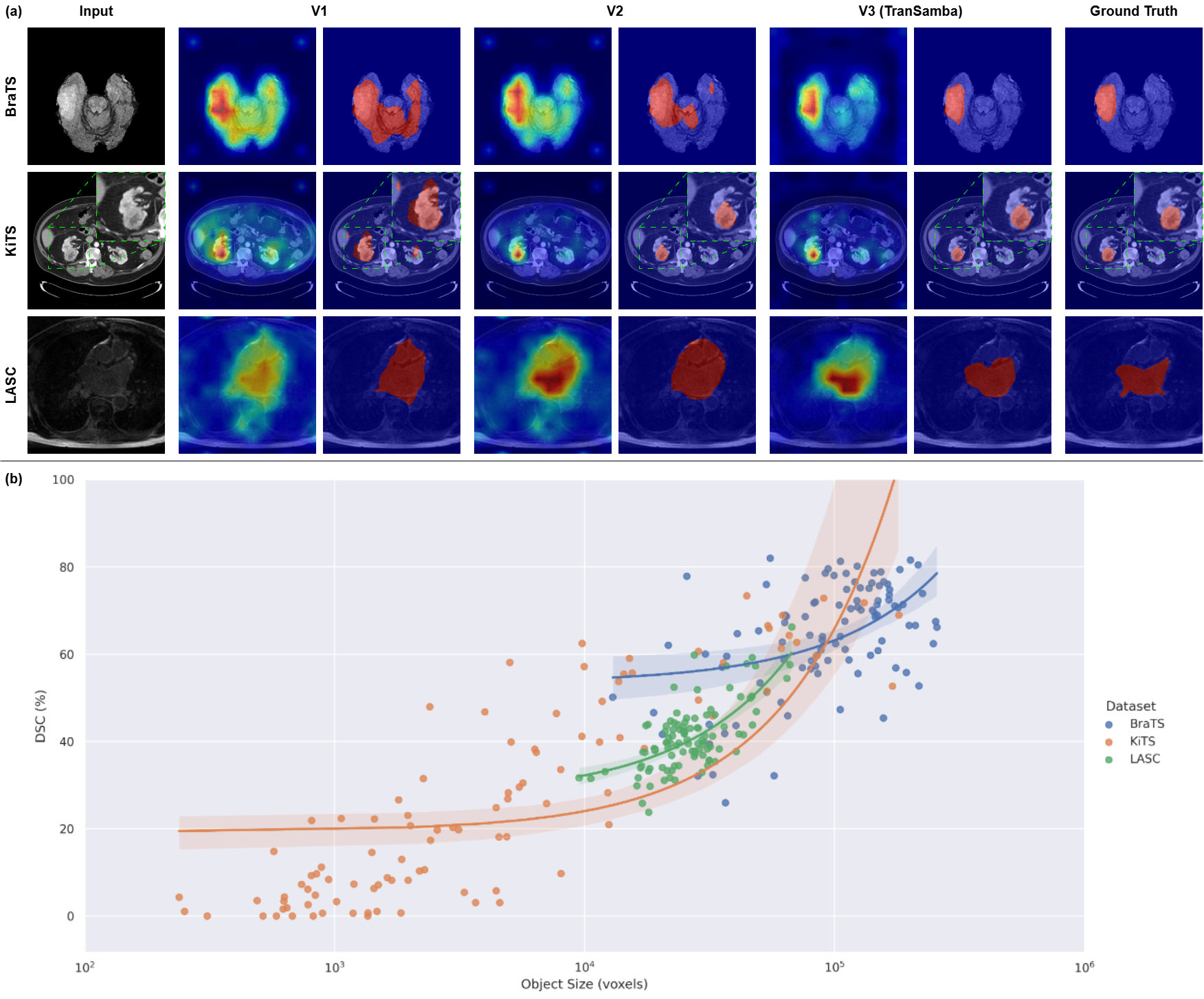}
\caption{\textbf{(a)} Qualitative comparison of V1--V3 (Tab.~\ref{tab2}(a)) on the validation set. \textbf{(b)} Size-DSC correlation of TranSamba on the testing set.} \label{fig2}
\end{figure}

\begin{table}[t]
\centering
\caption{Quantitative comparison with state-of-the-art methods on the testing set. Best and second-best results are highlighted in bold and italics, respectively. *: Hybrid backbone built on DeiT-S and vanilla Mamba.}\label{tab3}
\fontsize{8}{9.5}\selectfont
\begin{tabular}{cc|cc|cc|cc}
\hline
\multirow{2}{*}{Method} & \multirow{2}{*}{Backbone} & \multicolumn{2}{c|}{BraTS} & \multicolumn{2}{c|}{KiTS} & \multicolumn{2}{c}{LASC}\\
 & & DSC (\%) & IoU (\%) & DSC (\%) & IoU (\%) & DSC (\%) & IoU (\%)\\
\hline
Grad-CAM \cite{selvaraju2017grad} & DeiT-S & 55.2 & 36.6 & 10.7 & 15.0 & 6.0 & 2.8\\
Grad-CAM++ \cite{chattopadhay2018grad} & DeiT-S & 45.9 & 28.3 & 3.8 & 5.6 & 8.8 & 5.9\\
FullGrad \cite{srinivas2019full} & DeiT-S & 33.2 & 19.8 & 20.5 & \textit{25.1} & \textit{37.7} & \textit{23.1}\\
Ablation-CAM \cite{ramaswamy2020ablation} & DeiT-S & 52.5 & 34.8 & 7.4 & 9.4 & 5.6 & 2.6\\
Score-CAM \cite{wang2020score} & DeiT-S & 32.4 & 21.9 & 0.8 & 0.8 & 4.3 & 2.0\\
SEAM \cite{wang2020self} & ResNet-38 & 20.8 & 8.5 & 7.7 & 6.1 & 9.5 & 5.2\\
LayerCAM \cite{jiang2021layercam} & DeiT-S & 29.4 & 17.8 & \textit{21.4} & 17.3 & 0.3 & 0.2\\
TS-CAM \cite{gao2021ts} & DeiT-S & 50.6 & \textit{42.8} & 7.8 & 20.9 & 16.5 & 13.5\\
SIPE \cite{chen2022self} & ResNet-50 & 19.9 & 10.5 & 3.3 & 2.7 & 4.0 & 1.9\\
AME-CAM \cite{chen2023ame} & ResNet-18 & \textit{57.0} & 42.2 & 3.5 & 2.3 & 1.1 & 0.6\\
UM-CAM \cite{fu2023cam} & VGG-16 & 51.4 & 35.5 & 13.2 & 17.1 & 24.2 & 15.3\\
IAT \cite{lyu2024importance} & DeiT-S & 52.8 & 37.1 & 10.7 & 11.4 & 36.1 & 21.9\\
\hline
\rowcolor{gray!50} TranSamba & DeiT-S* & \textbf{64.1} & \textbf{46.8} & \textbf{26.8} & \textbf{25.8} & \textbf{41.0} & \textbf{25.6}\\
\hline
\end{tabular}
\end{table}

\subsection{Ablation Studies}
We ablate \textbf{1.} cross-plane modeling effectiveness \textbf{2.} varying plane numbers, and \textbf{3.} hybrid layer design.

\textbf{Effectiveness of Cross-Plane Modeling} TranSamba is compared with no-Mamba variants; Tab.~\ref{tab2}(a,b) presents the results. TranSamba (V3) outperforms V1, a ViT-only baseline, by large margins. To demonstrate this improvement is not from increased complexity, we introduce V2, also comprising alternating Transformer and in-plane Mamba blocks equivalent to the forward branch in Vision Mamba (ViM) \cite{zhu2024vision}. With comparable complexity to V3, V2 yields much lower improvement over V1, indicating the performance improvement of V3 primarily stems from cross-plane modeling. Fig.~\ref{fig2}(a) shows a qualitative comparison of V1--V3; V3 achieves the most conformal segmentation. For completeness, we report Mamba-only variant performance: V4 uses ViM-like bidirectional SSMs for in-plane modeling; V5 further augments V4 with CPM blocks; object localization relies on patch token-based CAM generation. Compared to V1--V3, V4 and V5 achieve substantially inferior performance, justifying using SA for in-plane modeling despite higher computational complexity; we acknowledge the lower speed of V4 and V5 results from implementation inefficiency. Nonetheless, V5 outperforms V4 by incorporating cross-plane modeling.

\textbf{Number of Planes}
We evaluate performance across varying plane numbers: the minimum is 16; the maximum cannot exceed the lowest dataset plane count (155/71/44 for BraTS/KiTS/LASC). As shown in Tab.~\ref{tab2}(c), modeling across 16 planes achieves the highest performance. The target objects are continuous structures spanning a limited number of planes; hence, local cross-plane modeling is more beneficial than long-range modeling.

\textbf{Hybrid Layer Design} Regardless of the design, cross-plane modeling consistently improves the ViT baseline. TranSamba adopts the CrossIn design, which achieves the tied-highest ranking of 1.83 (Tab.~\ref{tab2}(d)) and the lowest HD95 on KiTS and second-lowest HD95 on BraTS and LASC (BraTS/KiTS/LASC: \textit{4.6}/\textbf{15.2}/\textit{3.2}; Parallel: 4.8/15.5/\textbf{3.1}, InCross: \textbf{3.8}/\textit{15.4}/3.6). A CrossIn layer places the CPM block before the Transformer block, ensuring C2P attention extraction follows cross-plane modeling, thereby maximizing map quality improvement.

\subsection{Comparison with State-of-the-Art}
Tab.~\ref{tab3} compares TranSamba with SOTA methods based on image-/plane-level labels. TranSamba achieves the highest performance across all three datasets, outperforming the second-highest performer of each dataset by mean margins of 5.3\% and 2.4\% in DSC and IoU. Notably, AME-CAM \cite{chen2023ame} and UM-CAM \cite{fu2023cam} perform dynamic multi-scale feature aggregation, while IAT \cite{lyu2024importance} adopts focal loss and self-supervision for intra-class heterogeneity. These strategies are effective on BraTS; however, performance drops substantially on KiTS and LASC. In contrast, TranSamba demonstrates robustness across tasks, indicating the generalizable effectiveness of cross-plane modeling.

Most methods achieve higher performance on BraTS than the other datasets. While the BraTS background is normal brain tissue, KiTS and LASC backgrounds are noisy cavities; LASC also has a small training set (43 volumetric studies). Despite performance gains from our hybrid architecture, a limitation is that TranSamba requires post-processing or multi-step approaches for clinical applications.

\begin{table}[t]
\centering
\caption{Quantitative results of TranSamba grouped by object size on the testing set.}\label{tab4}
\fontsize{8}{9.5}\selectfont
\begin{tabular}{cc|cc|cc|cc}
\hline
\multirow{2}{*}{Subgroup} & Size & \multicolumn{2}{c|}{BraTS} & \multicolumn{2}{c|}{KiTS} & \multicolumn{2}{c}{LASC}\\
 & (Voxels) & \#studies & DSC (\%) & \#studies & DSC (\%) & \#studies & DSC (\%)\\
\hline
Tiny & $< 10^3$ & 0 & N/A & 23 & 5.1 & 0 & N/A\\
Small & $[10^3, 10^4)$ & 0 & N/A & 50 & 21.3 & 1 & 31.7\\
Large & $[10^4, 10^5)$ & 47 & 58.3 & 24 & 54.3 & 99 & 41.1\\
Huge & $\ge 10^5$ & 53 & 69.2 & 3 & 64.5 & 0 & N/A\\
\hline
\end{tabular}
\end{table}

\textbf{Object Size Analysis} To evaluate model sensitivity to object size, studies in the testing set are assigned to four subgroups according to target object size (in voxels): tiny, small, large, and huge (Tab.~\ref{tab4}). Fig.~\ref{fig2}(b) shows the size distribution and size-DSC correlation of TranSamba. BraTS and LASC objects have a more clustered distribution, while KiTS size distribution spans three orders of magnitude. Most KiTS objects are much smaller, resulting in the lower absolute DSC on KiTS. Within each dataset, TranSamba performance exhibits a positive correlation with object size. Tab.~\ref{tab4} shows performance by subgroup; TranSamba also achieves the highest DSC for each subgroup, with the exception of the huge subgroup of KiTS (FullGrad: 66.1\%). The size-DSC correlation indicates another TranSamba limitation: performance drop on small objects, a well-known issue for medical segmentation \cite{wang2025s3}.

\section{Conclusion}
This paper introduces TranSamba, a hybrid Transformer-Mamba architecture to bridge 3D modeling and the reliance on 2D encoders in weakly supervised volumetric medical segmentation. The integration of CPM blocks with a ViT encoder enables efficient cross-plane modeling, which improves in-plane SA for object localization. TranSamba maintains linear time complexity and constant space complexity with the input volume depth, and achieves SOTA performance on three datasets, demonstrating the generalizable effectiveness of cross-plane modeling across diverse volumetric medical segmentation tasks. Future work could extend TranSamba to multi-class scenarios and incorporate auxiliary techniques for more clinically oriented applications.

\bibliographystyle{splncs04}
\bibliography{mybibliography}
%




\end{document}